\documentclass[letterpaper]{article} % DO NOT CHANGE THIS
\usepackage{aaai23}
\usepackage{times}  % DO NOT CHANGE THIS
\usepackage{helvet}  % DO NOT CHANGE THIS
\usepackage{courier}  % DO NOT CHANGE THIS
\usepackage[hyphens]{url}  % DO NOT CHANGE THIS
\usepackage{graphicx} % DO NOT CHANGE THIS
\usepackage{natbib}  % DO NOT CHANGE THIS AND DO NOT ADD ANY OPTIONS TO IT
\usepackage{caption} % DO NOT CHANGE THIS AND DO NOT ADD ANY OPTIONS TO IT
\usepackage{algorithm}
\usepackage{algorithmic}
\usepackage{soul}

\usepackage{newfloat}
\usepackage{listings}
\DeclareCaptionStyle{ruled}{labelfont=normalfont,labelsep=colon,strut=off} % DO NOT CHANGE THIS
\floatstyle{ruled}
\newfloat{listing}{tb}{lst}{}
\floatname{listing}{Listing}
\usepackage{todonotes}

\usepackage{todonotes}

\usepackage{todonotes}

\usepackage{todonotes}

\usepackage{todonotes}

\title{Exploiting Explainability to Design Adversarial Attacks and\\ Evaluate Attack Resilience in Hate-Speech Detection Models}

\author{%
Pranath Reddy Kumbam$^{1}$ \quad Sohaib Uddin Syed$^{1}$ \quad Prashanth Thamminedi$^1$ \quad Suhas Harish$^1$ \quad Ian Perera$^2$ \quad Bonnie J. Dorr$^1$\\
$^1$University of Florida \quad $^2$Institute for Human and Machine Cognition\\
\{kumbam.pranath, sohaibuddinsyed ,pthamminedi, sganjalaguntehar\}@ufl.edu\\
iperera@ihmc.org \quad bonniejdorr@ufl.edu
}

\begin{document}
\maketitle
\begin{abstract}
The advent of social media has given rise to numerous ethical challenges, with hate speech among the most significant concerns. Researchers are attempting to tackle this problem by leveraging hate-speech detection and employing language models to automatically moderate content and promote civil discourse. Unfortunately, recent studies have revealed that hate-speech detection systems can be misled by adversarial attacks, raising concerns about their resilience. While previous research has separately addressed the robustness of these models under adversarial attacks and their interpretability, there has been no comprehensive study exploring their intersection. The novelty of our work lies in combining these two critical aspects, leveraging interpretability to identify potential vulnerabilities and enabling the design of targeted adversarial attacks. We present a comprehensive and comparative analysis of adversarial robustness exhibited by various hate-speech detection models. Our study evaluates the resilience of these models against adversarial attacks using explainability techniques. To gain insights into the models' decision-making processes, we employ the Local Interpretable Model-agnostic Explanations (LIME) framework. Based on the explainability results obtained by LIME, we devise and execute targeted attacks on the text by leveraging the TextAttack tool. Our findings enhance the understanding of the vulnerabilities and strengths exhibited by state-of-the-art hate-speech detection models. This work underscores the importance of incorporating explainability in the development and evaluation of such models to enhance their resilience against adversarial attacks. Ultimately, this work paves the way for creating more robust and reliable hate-speech detection systems, fostering safer online environments and promoting ethical discourse on social media platforms. 
\end{abstract}

\section{Introduction}

Due to the expanding influence of social media, it has become increasingly important to understand the nature of online exchanges and address discussions that contain offensive or hateful content. While prior work has focused on the efficacy of content moderation \cite{kumar2019content}, recent attention has shifted to automated mediation for promoting civil discourse instead of merely removing posts that include offensive language. Steps in this direction are only in the nascent stages \cite{kirk-etal-2022-handling}. A crucial prerequisite for ensuring the effectiveness of approaches to achieving this goal is the accurate detection of hate speech.

An interaction may be offensive if it touches upon sensitive elements such as race, color, gender, ethnicity, religion, etc. The problem of discerning whether a sentence includes hate speech has been explored using models such as BERT \cite{2018arXiv181004805D}, LSTM \cite{10.1162/neco.1997.9.8.1735} and CNN \cite{2014arXiv1408.5882K}.  
Automated moderation involves sophisticated models that analyze and classify content, based on predefined rules and guidelines \cite{Ye2023exploring,wang2022governing}.
However, these models underperform even with slight perturbations in the input \cite{2014arXiv1412.6572G,explainTextClassifier2022,ye2023noisyhate}.  

Consider these two sentences: \newline
\mbox{~~~~~}(1) Voldemort is a bad person\newline
\mbox{~~~~~}(2) Voldemort is not a good person. \newline
Sentence (1) and Sentence (2) convey a similar meaning, but the latter confuses hate-speech models because the full understanding of the sentence relies not just on the adjectives "good" or "bad." This example showcases how adversarial AI can seamlessly introduce noise by adding the word "not" before a positive adjective, successfully hiding a negative sentiment within a post.

Previous studies have explored adversarial robustness \cite{2021arXiv210907403H} and interpretability of machine learning models \cite{a15080291} independently. Our work brings these two vital aspects together, setting our approach apart from prior work. We argue that an understanding of how interpretability can be exploited for adversarial attacks is necessary for the development of more robust hate-speech detection systems. We delve into this issue, investigating the relationship between interpretability and accuracy under adversarial attacks. We hypothesize that enhancing the explainability of a model may inadvertently expose its inherent vulnerabilities, making it more susceptible to perturbations and, in turn, compromising its adversarial robustness. This nuanced understanding of the tradeoff between model explainability and adversarial resilience drives the design of our research and development of solutions.

Automated hate-speech detection models are designed to combat the spread of harmful content on social media platforms. Neural and transformer models (BERT, LSTM, and CNN) classify and flag instances of hate speech for downstream processing, yet their performance is suboptimal in the face of subtle or sophisticated instances of hate speech. 

The ability for adversarial attacks to undetectably manipulate the models' inputs raises concerns as this can undermine the effectiveness of these systems. Specifically, adversarial examples are crafted by introducing perturbations to the input data, exploiting the models' vulnerabilities and causing them to misclassify the input. These perturbations can significantly degrade the models' performance, even if they don't affect human perception of hate speech. 

A first step toward tackling this challenge is to understand the relationship between interpretability and prediction accuracy under adversarial conditions. Interpretability sheds light on the underlying factors that contribute to the models' predictions and is essential for identifying potential biases, flaws, and vulnerabilities. Deep learning techniques lack transparency, hindering trust and accountability \cite{2017arXiv170208608D}. Various techniques have thus been proposed to improve model interpretability, e.g., LIME \cite{ribeiro_why_2016}, SHAP \cite{lundberg_unified_2017}, CAM \cite{2015arXiv151204150Z}, among others.

Machine learning models are increasingly used for online hate-speech detection, yet they are vulnerable to adversarial attacks in the form of manipulated inputs that are designed to cause hate-speech detection models to misclassify the input.  Adversarial attacks may involve subtle modifications to the text, such as word order changes, synonym replacements, or the addition of irrelevant content. These modifications may not affect the human perception of hate speech, but they can considerably degrade the models' performance.

Model interpretability refers to the ability to explain and understand the model's decision-making process, which is important for various reasons: (1) \textbf{Debugging}: Understanding reasons behind incorrect predictions is valuable for troubleshooting and resolving the issue; (2) \textbf{Trust}: Users are more likely to trust a model if they understand how it works; and (3) \textbf{Accountability}: Model accountability hinges on being able to explain discriminatory predictions.

In hate-speech detection, interpretability helps identify potential biases, flaws, and vulnerabilities in the models. This information guides the development of more robust and resilient systems. For example, adjusting training data or incorporating new features can mitigate biased outcomes, highlighting the importance of enhancing interpretability for improving reliability, robustness, and accountability of hate-speech detection models.

Several techniques have been designed to enhance the interpretability of machine learning models. For example, LIME (Local Interpretable Model-agnostic Explanations) approximates the model's behavior around a specific input instance using a locally-linear surrogate model, enabling explanations of its predictions for that input instance.  

To enhance our understanding of the strengths and vulnerabilities of state-of-the-art hate-speech detection models, we employ LIME as a means to evaluate their interpretability. We argue for the inclusion of explainability in the development and evaluation of these models to bolster their resilience against adversarial attacks. Ultimately, our aim is to cultivate safer online environments and encourage ethical discussions on social media platforms.

\section{Related Work}
\label{Literature_Survey}
The growing online presence of modern societies has fostered a widespread culture of freedom of speech, but it has also brought challenges such as racism and propagation of hate speech \cite{williams2020hate}. Identifying hateful comments, posts and content is crucial for automating the promotion of civil discourse. Language models such as BERT, CNN and BiRNN have been successful at detecting and classifying hate speech \cite{mathew2021hatexplain} but are prone to adversarial attacks and perturbations \cite{zhang2020adversarial}. 

Hate-speech detection is important for online content mediation, yet even the most advanced models have significant limitations. Traditional evaluation of hate-speech detection models relies on their performance on a limited set of commonly used datasets \cite{niven-kao-2019-probing}. However, this approach has limitations. First, aggregate performance metrics, such as accuracy, offer limited insight into specific model weaknesses \cite{10.1371/journal.pone.0243300}. For example, a model that achieves high accuracy on a dataset of hate speech may still be biased against certain groups. 

Second, models may perform deceptively well on held-out test sets due to gaps and biases in the training data. A hate speech model trained on a dataset primarily targeting a specific community may mistakenly flag content mentioning that community as hate speech, even if it is not actually hateful \cite{park-etal-2018-reducing}. These limitations are particularly problematic for hate-speech detection since current hate-speech datasets differ in data source, sampling strategy, and annotation process. 

Additionally, these datasets are known to exhibit annotator and author biases. As a result, models trained on such datasets may be overly sensitive to lexical features such as group identifiers and often generalize poorly to other datasets \cite{kennedy-etal-2020-contextualizing}. Thus, evaluating models solely on their performance on current hate-speech datasets is incomplete and potentially misleading. More comprehensive evaluation methods are needed that take into account the specific limitations of hate-speech detection models.

While evaluating model performance is important, it is also crucial to make the model more interpretable and explainable to ensure fairness and transparency in the decision-making processes \cite{MatherPKCGWD22}. Interpretability and explainability in machine learning are vital for identifying model biases, flaws, and vulnerabilities \cite{lipton_mythos_2016, molnarInterpretable}. Various techniques have been introduced to enhance these attributes, including intrinsic methods, model-specific post-hoc methods, and model-agnostic post-hoc methods like LIME \cite{ribeiro_why_2016} and SHAP \cite{lundberg_unified_2017}. LIME, in particular, has been employed in diverse domains and tasks, including hate-speech detection, to reveal key features and patterns that impact model predictions \cite{beltagy_longformer_2020}.

Adversarial tools such as TextAttack \cite{morris2020textattack} have been used to understand and evaluate the susceptibility of models to adversarial vulnerabilities. These tools provide a framework for conducting adversarial attacks in order to investigate adversarial robustness using pre-trained models. Although \citet{mozafari2020bert} offer a thorough analysis of BERT's use in hate-speech detection through transfer learning, there remains a need for evaluating these models against a wide range of perturbations. In addition, a qualitative analysis of the relationship between explainability and adversarial robustness is needed.

\section{Toward Interpretable, Adversarially Robust Hate-Speech Detection}
\label{Proposed_Solution}

We adopt a systematic approach to evaluating both interpretability and adversarial robustness of state-of-the-art hate-speech detection models, e.g., BERT, LSTM, and CNN. We examine the relationship between  Degree of Explainability (DoE) and Adversarial Robustness (Ar) for these models and gain insights into their vulnerabilities and resilience under adversarial attacks. We employ Local Interpretable Model-agnostic Explanations (LIME), a widely-used post-hoc technique for assessing interpretability of hate-speech models.

We analyze the explanations generated by LIME for individual predictions made by the models, with the objective of uncovering critical features and patterns that contribute to their decision-making processes. This analysis provides valuable information about potential biases, limitations, and vulnerabilities in the models, which can be exploited by adversarial attacks.

Drawing from insights gained from our explainability analysis, we generate targeted adversarial perturbations to evaluate the models' robustness and resilience under adversarial conditions. These perturbations maintain the semantic meaning of the original inputs while inducing misclassifications in the models. To ensure a comprehensive evaluation of the models' adversarial robustness we employ the following perturbations: synonym substitution, character insertion or deletion, and paraphrasing. 

Once adversarial perturbations are generated, we evaluate the performance of the hate-speech detection models on the perturbed dataset. This evaluation involves measuring the models' accuracy and other relevant metrics under adversarial conditions, providing a quantitative assessment of their robustness and resilience against adversarial attacks.

Lastly, we undertake an in-depth investigation of the association between the Degree of Explainability (DoE) and Adversarial Robustness (Ar) for the various models. This examination entails plotting the ratio of DoE to Ar and scrutinizing the emerging trends and patterns. By contrasting the models' performance in terms of interpretability and robustness, we pinpoint potential tradeoffs and synergies between these attributes and offer guidance for creating more robust and interpretable hate-speech detection systems.

Our approach provides a comprehensive and systematic method to assess the interpretability and adversarial robustness of state-of-the-art hate-speech detection models. By combining explainability analysis, adversarial perturbation creation, and model evaluation under adversarial conditions, we aim to gain a deeper understanding of the challenges and potential in developing reliable hate-speech detection systems. In the following sections, we describe the dataset and pre-processing procedures, outline the model training process, and present the results and analysis of our experiments.

\section{Datasets and Pre-processing}
\label{Data}

We leverage two distinct datasets that contain information about the prevalence of hate speech on social media platforms. These datasets offer a wide-ranging and representative selection of hate speech and offensive language cases, allowing for a thorough assessment of the effectiveness and resilience of the hate-speech identification models. The datasets have been gathered from separate sources and encompass diverse content to guarantee the models' ability to generalize across various contexts and scenarios.

\subsection{Hate Speech and Offensive Language}

Kaggle's Hate Speech and Offensive Language dataset \cite{samoshyn2020hate} focuses on hate speech and offensive language usage on Twitter. This dataset possesses several key attributes: (a) it includes the tweet text, representing the original, unprocessed textual data from Twitter; (b) it is structured for multi-class classification, with tweets categorized into ``Hate speech,'' ``Offensive language,'' and ``Neutral;'' (c) each record contains a class variable, indicating the class to which the tweet belongs; and (d) its 24,783 unique entries make it an abundant resource for training and evaluating our models. We leverage the multi-class structure to assess models' performance across a spectrum of offensive content, spanning from blatant hate speech to subtler forms of offensive language.

\subsubsection{UC Berkeley Measuring Hate-speech}

The  publicly released Berkeley Measuring Hate-speech dataset, used in an experiment for hate-speech detection in prior work \cite{kennedy2020constructing}, has several important features that complement those of the first dataset: (a) it comprises over 135,556 combined rows of data, with 39,565 unique comments that have been annotated with ordinal labels by 7,912 annotators; (b) it is designed for binary classification, with only two categories: ``Hate speech" or "Not hate speech;'' (c) the dataset's main outcome variable is the hate-speech score, ranging from 0 to 1, where $>$0.5 denotes hate speech, and $\leq$0.5 denotes \textit{not} hate speech. The binary classification enables a focused evaluation of models' performance in accurately detecting and classifying hate speech instances, simplifying our assessment. Each is described, in turn, below.

\subsection{Pre-processing}
To ensure the quality and consistency of the data, we perform several pre-processing steps on both datasets. These steps include converting the tweets to lowercase, removing punctuation, extra spaces, URLs, mentions, and hashtags. Tokenization is performed using the NLTK library for the CNN and LSTM models. For the implementation of \mbox{DistilBERT}, we use the DistilBertTokenizer from the HuggingFace transformers library to tokenize the text data.

Following tokenization, we perform lemmatization to convert the words to their base forms, which aids in reducing dimensionality and improving the generalizability of the models. To further reduce noise and focus on meaningful words, we remove stop words in tweets. Additionally, we balance the class distribution in the datasets to mitigate the effects of class imbalance on model training and evaluation.

These pre-processing steps ensure that the input data for the hate-speech detection models are clean, consistent, and representative of the underlying language patterns and structures. By carefully preparing the data and addressing potential issues such as class imbalance, we aim to create a robust foundation for training and evaluating the models, allowing for a more reliable and accurate assessment of their performance under adversarial conditions. 

\section{Model Training}
\label{Models}

This section presents the technical details behind the deep learning models (LSTM, CNN, and \mbox{DistilBERT}) we have implemented for the purpose of detecting hate speech. Before testing the deep learning models, we establish a baseline score using a Random Forest (RF) model \cite{598994,10.1023/A:1010933404324} during preliminary analysis.

The RF model is an ensemble learning model consisting of multiple decision trees, primarily employed for classification tasks. The construction of each tree is independent of the others, and at each node, a random subset of features is taken into account for splitting. The best split is determined based on a predefined criterion, and the tree grows until it reaches a maximum depth or the number of samples in a leaf node falls below a certain threshold. For our model, the number of trees in the forest is set to 10.

\subsection{LSTM}

We have developed and trained a Long Short Term Memory (LSTM) model \cite{10.1162/neco.1997.9.8.1735} to classify the text for hate-speech detection. LSTMs are a type of recurrent neural network that captures long-range dependencies in text, which makes them highly suitable for this particular text classification task. 

The model consists of three components, an embedding layer, an LSTM layer, and finally a fully connected layer. The embedding layer converts the input text into a dense vector representation, while the LSTM layer captures long-range dependencies within the text. Finally, the fully connected layer produces class predictions.

To train the model, we employ a supervised learning approach with the Adam optimizer \cite{2014arXiv1412.6980K} and the Cross-Entropy loss function. During the training process, the model’s parameters are iteratively updated using mini-batch gradient descent for a fixed number of epochs. To assess the model's performance we monitor the average training loss per epoch.

\subsection{CNN}

We have also used a convolutional neural network (CNN) \cite{article} to classify the text for hate-speech detection. The CNN model comprises an embedding layer, three convolutional layers with varying kernel sizes, and a fully connected layer.

The embedding layer is responsible for converting the words into vectors, capturing the semantic relationships between them. The convolutional layers are designed to capture local patterns or features within the input text. To capture different levels of abstraction in the text data, we have employed three kernel sizes: 3, 4, and 5. 

To reduce the spatial dimension of the feature maps, we have incorporated pooling layers into our CNN architecture. Finally, the fully connected layer maps the extracted features to the respective output classes, enabling text classification. 

\subsection{BERT}

We have additionally utilized the \mbox{DistilBERT} model \cite{2019arXiv191001108S} which is a lighter version of BERT \cite{2018arXiv181004805D} for this task. \mbox{DistilBERT} retains most of BERT’s performance while reducing the model’s size and computation cost, making it suitable for real-world applications. 
DistilBERT is a transformer-based model that leverages the self-attention mechanism to process the input text. The model’s architecture consists of an embedding layer followed by multiple transformer layers and a classification head. 

We use a pre-trained checkpoint from the HuggingFace transformers library as a starting point for model training. This checkpoint has undergone training and fine-tuning on our hate-speech dataset. We employ the AdamW optimizer \cite{2017arXiv171105101L} for model training, with a learning rate of $2e^{-5}$. To further optimize the training process, we apply a linear learning rate scheduler with warmup. The training process involves iterative updates to the model’s parameters using mini-batch gradient descent for a fixed number of epochs. The model’s performance is monitored using the average training loss per epoch.

\section{Determining Explainability \\for Hate-Speech Detection}
\label{Exp}

Our experimentation is designed to assess the explainability of the hate-speech detection models we have trained (LSTM, CNN, and \mbox{DistilBERT}). The goal is to gain a deeper understanding of how these models make decisions and provide insights into their decision-making processes. Through this evaluation, we expect to gain valuable insights into the inner workings of these models and identify potential areas for improvement. We employ the LIME technique, which which aids in interpreting and explaining the models' predictions.

\subsection{LIME}

Our research utilizes LIME (Local Interpretable Model-agnostic Explanations), a powerful technique that provides explanations for the predictions made by our hate-speech detection models, regardless of their underlying architecture. LIME generates localized explanations for individual predictions by approximating the model's behavior in the vicinity of a particular instance using simpler, interpretable model, such as a linear model. 

First, we select an instance for which an explanation is desired. Next, we perturb the selected instance and obtain predictions from the model for the perturbed instances. The perturbed instances are assigned weights based on their similarity to the original instance. A simpler, interpretable model is trained using the perturbed instances. During this training process, the model's predictions serve as targets, while the instance weights act as sample weights. Finally, the simpler model is interpreted to gain insights into the model's behavior specifically around the chosen instance.

Applying LIME to our hate-speech detection models for a given sentence provides an output in the form of $(W, <score>)$, where $W$ is a word in the sentence, and $<score>$ represents the prediction probabilities for that word. The score indicates the contribution of the word W towards the labeling of the sentence made by the model. By examining these word-level explanations, we acquire valuable insights into the features and patterns that the models rely on for their predictions. 

Through the use of the LIME technique, we aim to gain a comprehensive understanding of the decision-making processeses employed of our trained hate-speech detection models, while evaluating their explainability. This knowledge plays a crucial role in establishing trust in the models' predictions and ensuring that they are both accurate and interpretable. This is particularly important in sensitive domains such as hate-speech detection, where transparency and reliability are crucial factors.

\section{Adversarial Attacks}
\label{Adv}

After having assessed the explainability of each hate-speech detection model (LSTM, CNN, and \mbox{DistilBERT}) using our experimentation process, we move on to the introduction of adversarial attacks. The goal of these attacks is to modify a portion of the original text while preserving the intended meaning of the sentence. To execute adversarial attacks, we utilize the TextAttack library, which requires inputs such as the sentence to be attacked, the associated label (hate-speech or non-hate-speech), the model, and the tokenizer.

\subsection{Perturbations with Text Attack}

TextAttack is a versatile tool for generating adversarial examples in natural language processing tasks. It leverages the results of model explainability to identify words that have high explainability scores and replaces them with synonyms. The goal is to maintain the overall meaning of the sentence while potentially altering the model's classification.

The results obtained from TextAttack can be categorized into three types: successful, failed, and skipped. Successful results occur when TextAttack finds a suitable synonym and successfully changes the classification of the sentence (e.g., from hate-speech to non-hate-speech, or vice versa). Failed results arise when TextAttack identifies a synonym but fails to alter the classification. Skipped results happen when the tool is unable to find an appropriate synonym or encounters other issues during the attack process.

For the successful results, TextAttack provides the original sentence, the modified sentence, and the classification probabilities for both hate-speech and non-hate-speech categories. LIME is then applied to the modified sentences to compare the explainability results before and after the attack. The observed variations in the explainability results indicate the effectiveness of the adversarial attack.

\begin{figure}[t] 
\begin{small}
	\centering
	\includegraphics[width = 3.3in]{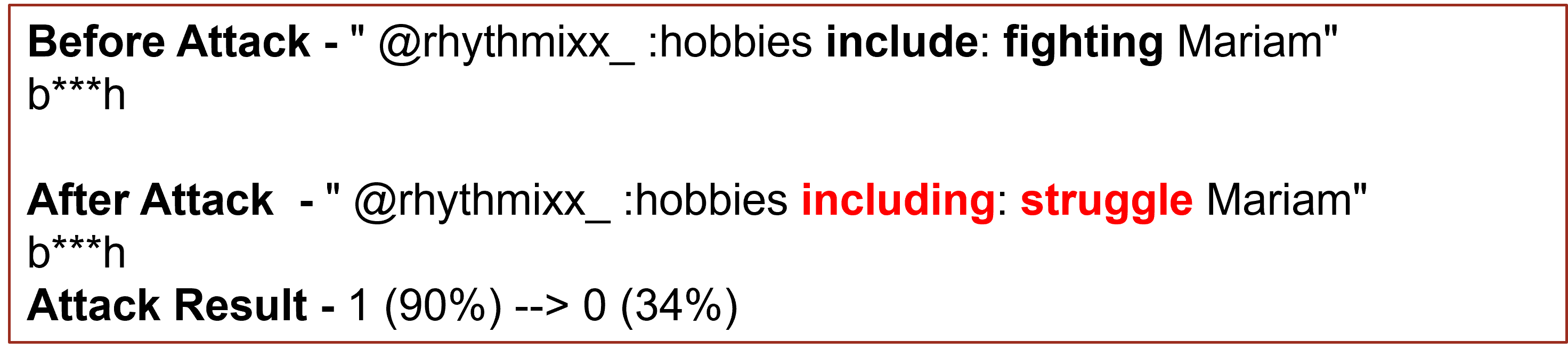}
	\caption{Text Attack 1}	
  \label{fig:Fig1}
\end{small}
\end{figure}
\begin{figure}[h] 
\begin{small}
	\includegraphics[width = 3.3in]{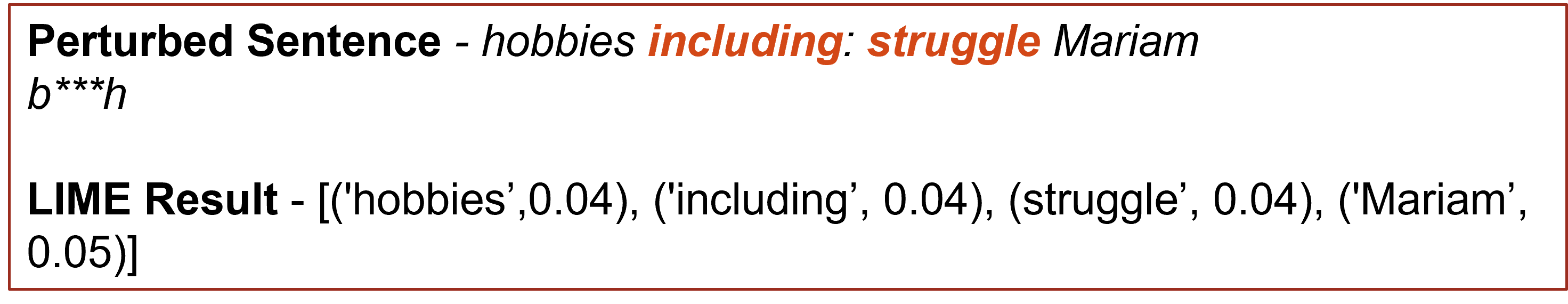}
        \caption{LIME Results of Text Attack 1}	
  \label{fig:Fig2}
\end{small}
\end{figure}

Figure \ref{fig:Fig1} illustrates a successful attack where a minor modification to the sentence leads the LSTM model to misclassify the input as neither offensive nor hate-speech, despite the appearance of an derogatory term (b***h).

%\hl{[BJD: I added the ``despite'' clause above; if you are fine with it, delete this comment -- if not please modify.]}

Figure \ref{fig:Fig2} demonstrates the deviation in LIME results between the original and modified sentences, highlighting the increased susceptibility of the sentence to adversarial attacks. The low scores assigned by LIME to the words in the perturbed sentence indicate that none of these words significantly contributes to classifying the sentence as hate speech, reflecting a successful adversarial attack.

%\hl{[BJD: I'm having trouble fully understanding the point here--are you saying that the scores are so low in the LIME Result that there is nothing to flag this as hate speech?  If so, please be clear here by adding a sentence or two.]}

\begin{figure}[h] 
\begin{small}
	\includegraphics[width = 3.3in]{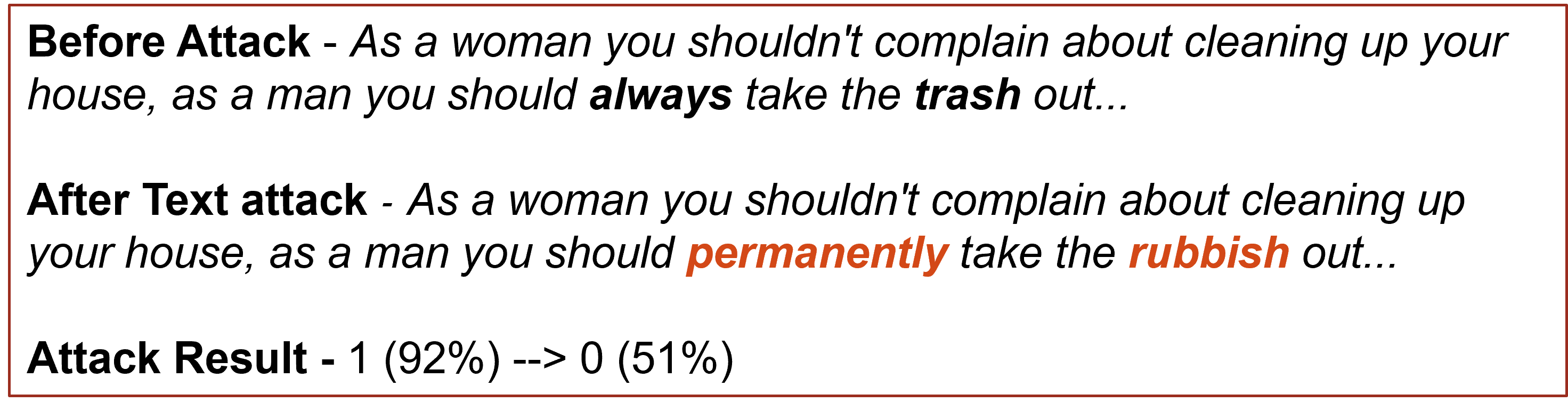}
        \caption{Text Attack 2}	
  \label{fig:Fig3}
  \end{small}
\end{figure}

Figure \ref{fig:Fig3} presents another example where a slight change in the sentence causes the model to classify it as non-offensive. This sentence promotes damaging gender stereotypes, suggesting that it is offensive and should be categorized as such.
%\hl{[BJD: Please add a sentence here stating that this is indeed offensive, and state why, to make it clear.]}

Through the use of adversarial attacks and the TextAttack tool, we assess the robustness of our hate-speech detection models against malicious manipulations. By identifying weaknesses and improving the models, we can enhance their resilience against adversarial attacks and ensure their reliability in real-world applications.

\section{Metrics and Evaluation}
\label{Metrics}

Evaluating the performance of hate-speech detection models is crucial to ensuring their reliability and effectiveness in real-world applications. By obtaining the explainability results from LIME and the attack outcomes from the TextAttack tool for various models, we establish a comprehensive framework to assess their Adversarial Robustness. We employ the following metrics to gauge their performance:

\subsection{Degree of Explainability}

LIME offers a valuable tool for assessing the importance of individual words in a sentence when it comes to the classification decisions made by the model. By providing a dictionary of values, known as prediction probabilities, we assess the impact of each word in determining whether a sentence qualifies as hate speech or not. 

To quantify the Degree of Explainability, we first calculate the standard deviation of the prediction probabilities for each sentence. Next, we determine the fraction of words whose prediction probability surpasses the standard deviation. A high Degree of Explainability indicates that the model's classification decision is primarily driven by a few words with substantial explainability scores. As a result, altering these words could substantially change the explainability outcome, rendering the sentence more vulnerable to adversarial attacks.

\subsection{Adversarial Robustness}

To measure a model's resilience against adversarial attacks, we quantify the Adversarial Robustness by computing the ratio between the model's accuracy under attack and its accuracy before the attack. A ratio near 1 signifies a high level of robustness, indicating that the model can withstand adversarial manipulations. Conversely, a lower ratio suggests an increased vulnerability to adversarial attacks.

\subsection{Attack Resilience}

Understanding the relationship between Adversarial Robustness and Degree of Explainability is crucial for developing more effective and reliable hate-speech detection models. We compute the ratio of Adversarial Robustness to Degree of Explainability, which we define as Attack Resilience, to assess this relationship.

A high Degree of Explainability suggests that a model is easily interpretable, but it also makes it more susceptible to adversarial attacks, resulting in lower Adversarial Robustness and Attack Resilience. On the other hand, a low Degree of Explainability means that the model is less interpretable but more resistant to adversarial attacks, yielding higher Attack Resilience. The ideal approach is to find a balance between interpretability and resilience in order to achieve optimal Attack Resilience.

By employing these metrics, we can evaluate the performance of various hate-speech detection models, identify areas for improvement, and guide the development of more robust models for real-world applications. This comprehensive analysis enables researchers and practitioners to make informed decisions about the design and deployment of hate-speech detection models. This, in turn, facilitates the development of more effective solutions for addressing online hate speech while ensuring their robustness against adversarial perturbations.

\section{Results}
\label{Results}

We evaluate the performance of each language model, \mbox{DistilBERT}, CNN, LSTM, as well as a baseline Random Forest Classifier (RFC). We use Accuracy (A), Precision (P), Recall (R), F1, and Area Under the Receiver Operating Characteristic Curve or AUC-ROC (abbreviated AUC in our results) to assess the models on two datasets. These metrics provide a comprehensive assessment of each model’s ability to detect hate speech and offensive language. The obtained results are used to compare the performance of each model against the others.

Table \ref{table:performance_comparison_kaggle} presents our experiment results for multi-class classification on Kaggle's multi-class ``Hate Speech and Offensive Language Datset.''  Table \ref{table:performance_comparison_ucb} presents the corresponding results for binary classification, on the ``Measuring Hate-Speech Dataset.'' 

\begin{table}[h]
\centering
\begin{small}
\begin{tabular}{|l|l|l|l|l|l|}
\hline
\textbf{Model} & \textbf{A} & \textbf{P} & \textbf{R} & \textbf{F1} & \textbf{AUC} \\ \hline
DistilBERT                             & 0.91  & 0.91   & 0.91 & 0.91 & 0.94 \\ \hline
RFC               & 0.88  & 0.87    & 0.88 & 0.87 & 0.88 \\ \hline
CNN                                     & 0.88  & 0.86   & 0.88 & 0.87 & 0.91 \\ \hline
LSTM                                    & 0.87  & 0.85   & 0.87 & 0.86 & 0.88 \\ \hline
%RFC (Baseline)     & 0.83   & 0.83   & 0.83  & 0.82 & 0.86 \\ \hline
\end{tabular}
\caption{Comparison of Model Performance on Kaggle's multi-class ``Hate Speech and Offensive Language Dataset.''}
\label{table:performance_comparison_kaggle}
\end{small}
\end{table}

\begin{table}[h]
\centering
\begin{small}
\begin{tabular}{|l|l|l|l|l|l|}
\hline
\textbf{Model} & \textbf{A} & \textbf{P} & \textbf{R} & \textbf{F1} & \textbf{AUC} \\ \hline
DistilBERT                            & 0.97  & 0.97   & 0.97 & 0.97 & 0.99 \\ \hline
LSTM                                    & 0.96  & 0.96   & 0.96 & 0.96 & 0.99 \\ \hline
CNN                                     & 0.96  & 0.96   & 0.96 & 0.96 & 0.98 \\ \hline
RCF                & 0.95  & 0.95    & 0.95 & 0.95 & 0.98 \\ \hline
\end{tabular}
\end{small}
\caption{Comparison of Model Performance on UC Berkeley's binary class ``Measuring Hate-Speech Dataset''}
\label{table:performance_comparison_ucb}
\end{table}

These results clearly indicate that the \mbox{DistilBERT} model outperforms the other models across all evaluation metrics. It achieves the highest accuracy, precision, recall, and F1-score in both datasets. This suggests that \mbox{DistilBERT} is the most suitable model for hate-speech detection among the models tested. 
\begin{figure}[h] 
	\centering
	\includegraphics[width = 3.3in]{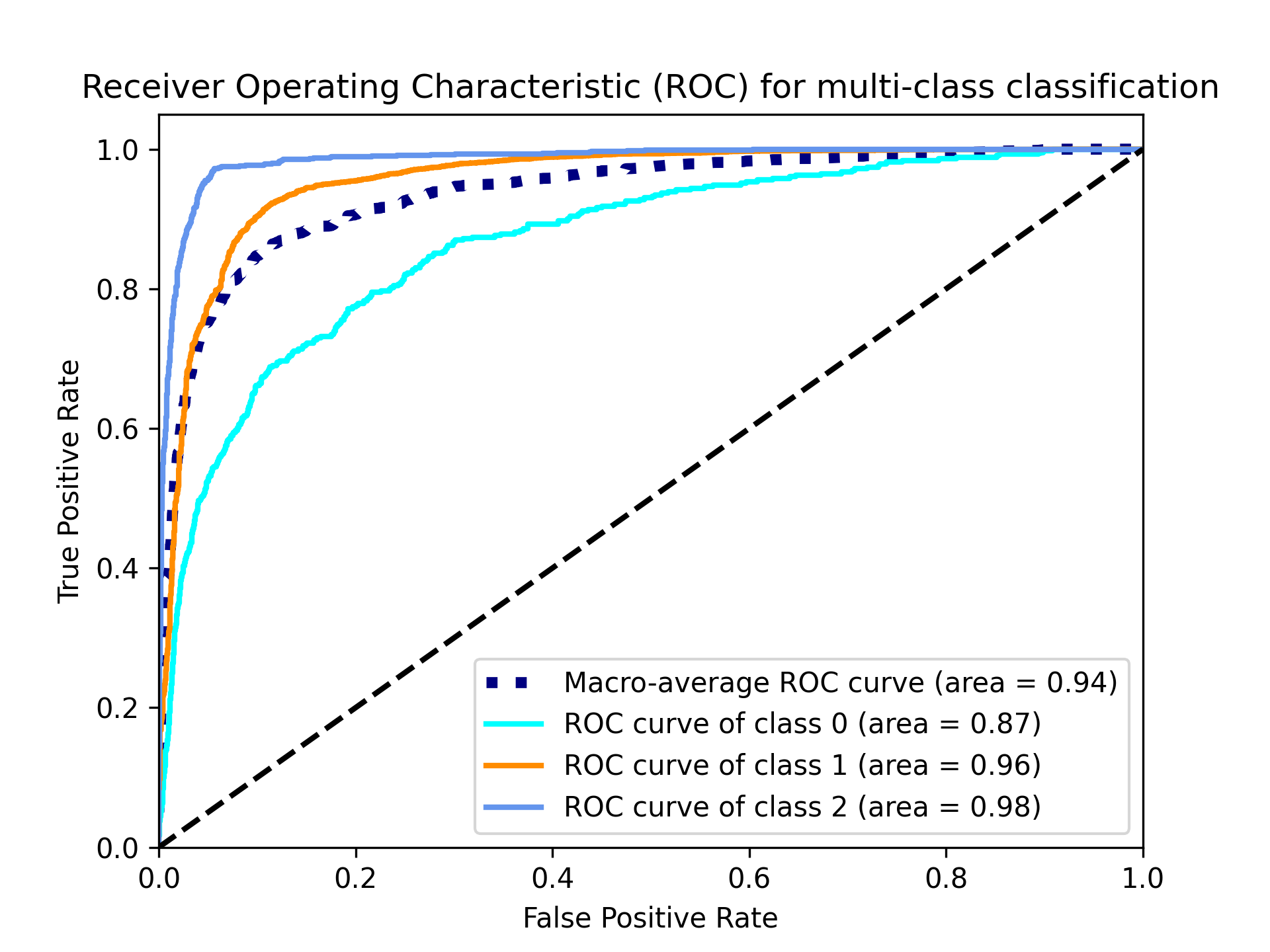}
	\caption{DistilBERT AUC-ROC Curve}	
  \label{fig:Fig4}
\end{figure}  

Additionally, Figure~\ref{fig:Fig4} provides the AUC-ROC curve for multi-class classification using the \mbox{DistilBERT} model as a reference. This curve reveals interesting insights related to the class imbalance in our dataset. Specifically, the AUC for class 2 (neutral) is greater than that for class 1 (offensive language), which in turn is greater than that for class 0 (hate speech). This pattern could likely be attributed to the class imbalance present in our dataset. As class 2 has a higher representation in the dataset, the model tends to perform better in correctly identifying these instances, resulting in a higher AUC. This is an important observation that shows the effect of class imbalance on our model's discriminative ability. This correlation between AUC scores and class representation underscores the need to address class imbalance in future model development.

%\hl{A sentence or two is needed here to highlight your key point about the ROC curve, please?  Or give just a brief remark here, and leave a forwarding pointer to later material.}

\subsection{BERT Analysis}
BERT (Bidirectional Encoder Representations from Transformers) is a powerful language model that demonstrates exceptional performance in various NLP tasks due to its effective handling of several key language phenomena. By employing bidirectional context processing, BERT supports a deeper understanding of the language and enhances the accuracy of its predictions. Furthermore, its pre-training involves exposure to large corpora through unsupervised tasks such as masked language modeling (MLM) and next sentence prediction (NSP), thus fostering substantial language understanding prior to fine-tuning for specific tasks. 

Built on the Transformer architecture, BERT leverages self-attention mechanisms to capture long-range dependencies and model complex relationships between words. Thus, it excels at handling hierarchical and contextual information. BERT also leverages transfer learning by fine-tuning the pre-trained model for specific tasks. This approach reduces the amount of labeled data and training time required for improved performance. 

Finally, BERT employs WordPiece tokenization to effectively handle out-of-vocabulary and rare words, thereby enhancing its capacity to learn meaningful representations. This capability significantly contributes to BERT's performance in tasks such as hate-speech detection through classification. By addressing these key language phenomena, BERT, and by extension, \mbox{DistilBERT}, demonstrate superior performance in hate-speech detection compared to other approaches that might not effectively handle these aspects of natural language.

For the multi-class classification task, the Random Forest Classifier, CNN, and LSTM models show competitive performance, with the Random Forest Classifier being the second-best performer. However, all three models have significantly lower performance compared to \mbox{DistilBERT}. It is worth noting that deep learning models, such as CNN and LSTM, typically require more training data and computational power to achieve better results. 
% Considering that these models are outperformed by the Random Forest Classifier, this suggests that the models have not converged and may require more data and/or training epochs. 
% The baseline Random Forest Classifier, on the other hand, demonstrates the lowest performance among all tested models in the multi-class classification task, owing to the difference in the data pre-processing techniques used. 

% In the binary classification task, the performance gap between \mbox{DistilBERT} and the other models (LSTM and CNN) is less significant. Both the LSTM and CNN models show comparable performance across all evaluation metrics, with only a marginal difference from \mbox{DistilBERT}. This suggests that, in a binary classification context, these models might also be considered for hate-speech detection, depending on the specific requirements and constraints. Considering the equivalent performance of the models, this dataset and task might be a more ideal choice for comparing the explainability and adversarial robustness of the models. 

In addition to the aforementioned evaluation metrics, we also calculate the Area Under the Receiver Operating Characteristic Curve (AUC-ROC) scores for both multi-class and binary classification tasks. 
% The AUC-ROC is a widely used and significant metric in machine learning, as it provides a comprehensive assessment of the classifier's performance across varying thresholds. By considering the true positive rate (sensitivity) and false positive rate (specificity) at different thresholds, the AUC-ROC score is capable of measuring the classifier's ability to distinguish between classes effectively. The inclusion of AUC-ROC scores in our analysis allows us to further evaluate the models' performance in detecting hate speech and offensive language, enabling a more robust comparison between the models. As opposed to the other metrics like accuracy, precision, and recall, which depend on a specific threshold, the AUC-ROC score represents the model's overall performance across a range of decision boundaries. This aspect is particularly important in the context of hate-speech detection, where the optimal threshold may vary depending on the specific application or platform.
Our results indicate that the AUC-ROC scores for multi-class classification follow a similar trend to the other evaluation metrics, with the \mbox{DistilBERT} model outperforming the other models. 
% nterestingly, the AUC-ROC score for the CNN model is higher than the Random Forest Classifier in the multi-class setting. This suggests that, despite the lower performance in terms of accuracy, precision, recall, and F1-score, the CNN model may have a better ability to discriminate between classes at varying thresholds, which could be an important consideration in certain use cases. In the binary classification task, the AUC-ROC scores also align with the overall trend observed in the other evaluation metrics. Although the performance gap between \mbox{DistilBERT} and the other models is much less significant, all models achieve near-unity AUC scores, indicating that they exhibit excellent performance in distinguishing between the two classes.

The inclusion of AUC-ROC scores in our evaluation provides a more comprehensive assessment of the model's performance for hate-speech detection. The AUC-ROC scores not only confirm the superior performance of the \mbox{DistilBERT} model but also highlight the potential strengths and weaknesses of the other models in differentiating between classes across various thresholds. This information is essential for informed decision-making when selecting the most suitable model for a particular hate-speech detection application or platform.

\subsection{Adversarial Attacks and Explainability}
\begin{figure}[h] 
	\centering
	\includegraphics[width = 3.3in]{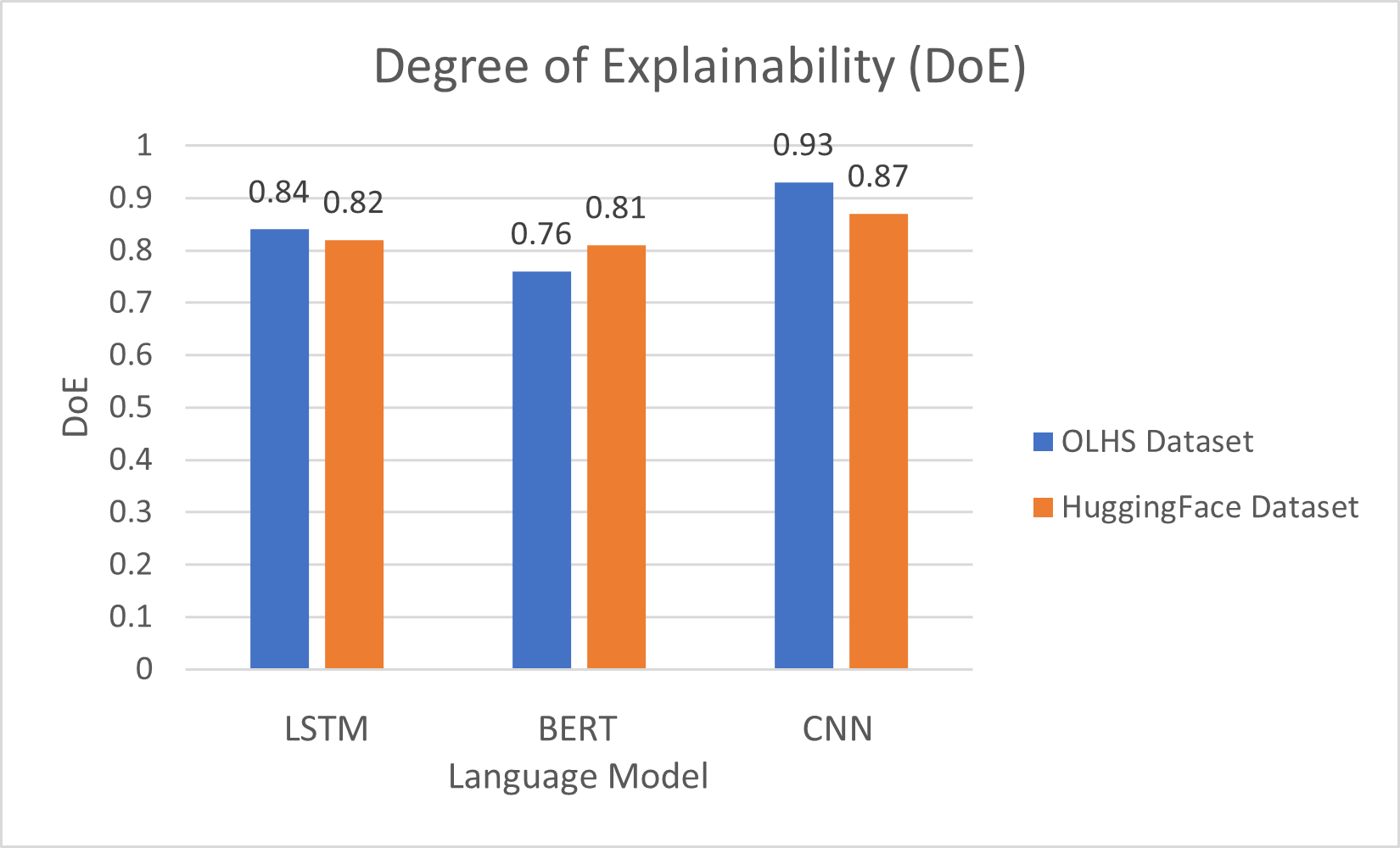}
	\caption{Degree of Explainability - LSTM, BERT and CNN}	
  \label{fig:DoE}
\end{figure}

Following experimentation, we have undertaken additional analysis to explore the tradeoff between the degree of explainability (DoE) and adversarial robustness. Figure \ref{fig:DoE} illustrates our results for degree of explainability. Our hypothesis suggests that models with higher DoE, such as CNN, should exhibit lower robustness to adversarial attacks. We will see (shortly) that this hypothesis is validated.

\begin{figure}[h] 
	\centering
	\includegraphics[width = 3.3in]{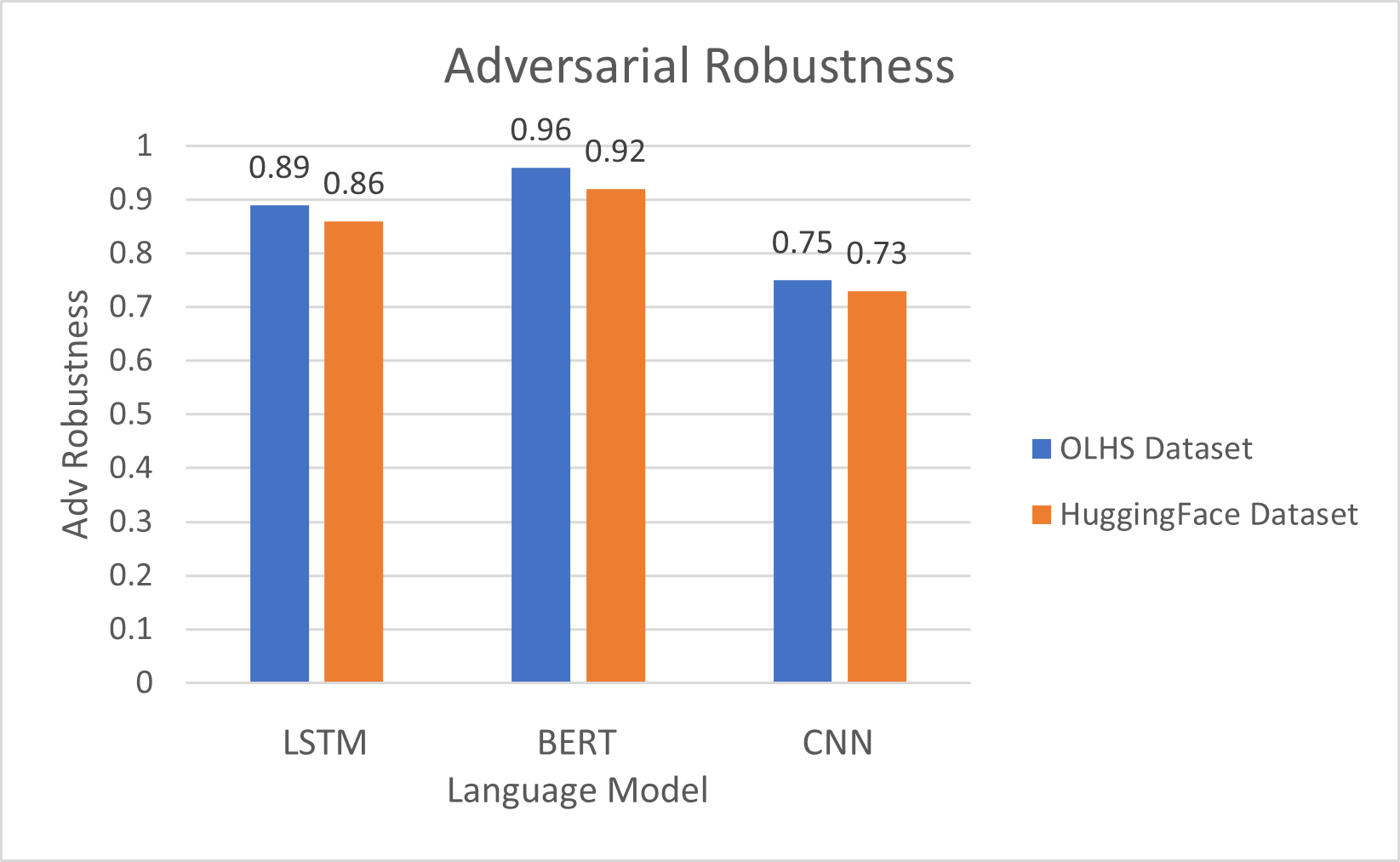}
	\caption{Adversarial Robustness - LSTM, BERT and CNN}	
  \label{fig:AdvRob}
\end{figure}

On the flipside, we expect that models with lower DoE, such as BERT, would demonstrate greater robustness. This expectation is validated by our results, as shown in Figure~\ref{fig:AdvRob}. BERT exhibits higher adversarial robustness compared to LSTM, which, in turn, surpasses CNN in terms of robustness. This result closes the loop on the hypothesized DoE/Robustness tradeoff pointed out above.

\begin{figure}[h] 
	\centering
	\includegraphics[width = 3.3in]{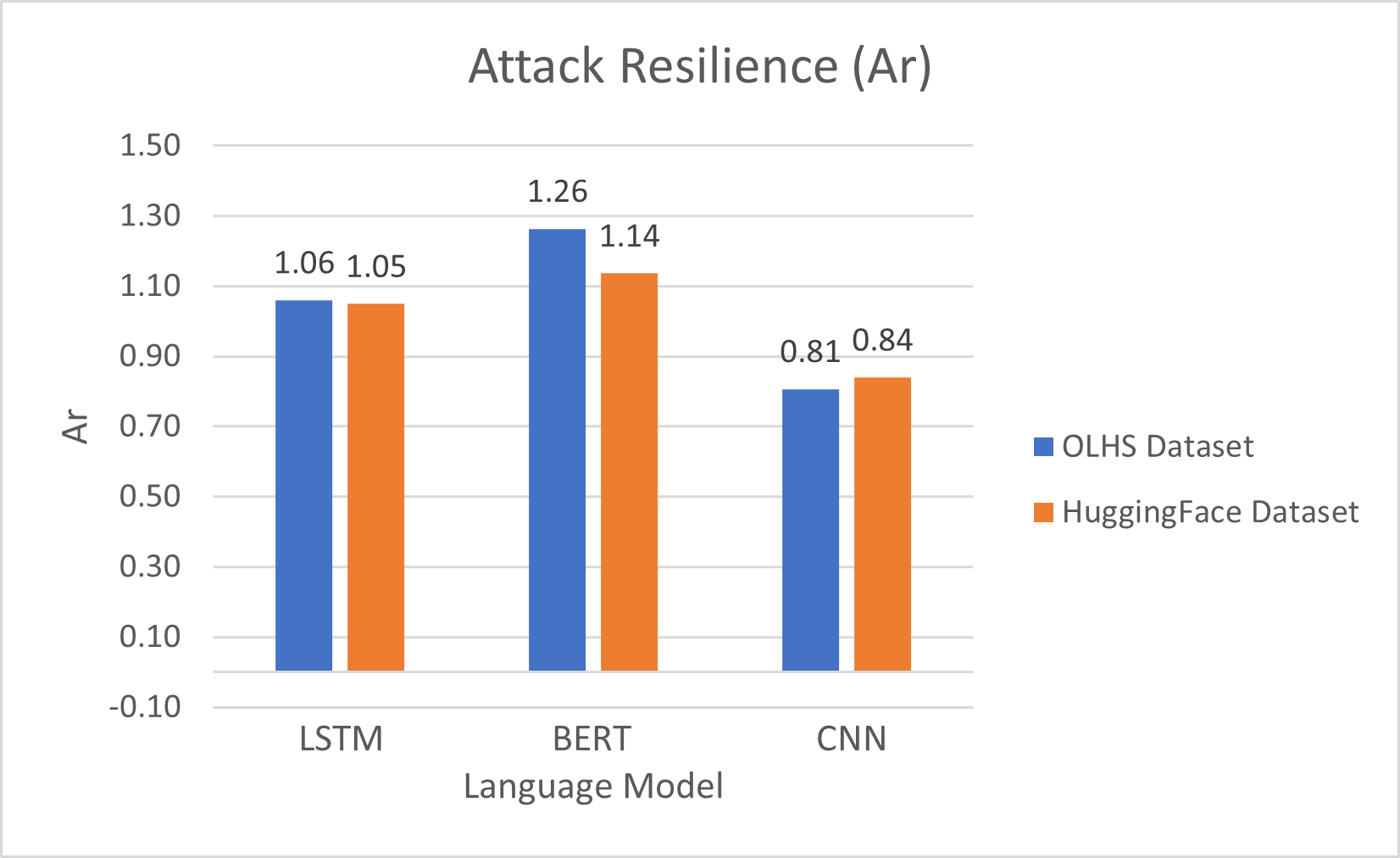}
	\caption{Attack Resilience - LSTM, BERT and CNN}	
  \label{fig:AR}
\end{figure}

Figure \ref{fig:AR} plots the \textit{attack resilience} which is the ratio of DoE to Adversarial Robustness. We see here that the BERT model exhibits the highest resilience to adversarial attacks among the three models, while the CNN model is the most vulnerable. 

Various factors may contribute to the differences in adversarial robustness shown above. BERT, a transformer-based model, demonstrates higher robustness due to its attention mechanism, which enables better capturing of long-range dependencies and contextual information more effectively than other models. Additionally, BERT's pre-training and fine-tuning processes may also contribute to its resilience, as they involve extensive learning from diverse data improved generalization to different inputs.

In contrast, LSTMs, as recurrent neural networks, can capture some sequential information and dependencies, but their memory capacity is relatively limited compared to the attention mechanism employed by BERT. As a result, LSTMs exhibit moderate robustness but are not as resilient as BERT. CNNs, designed primarily for handling grid-like data such as images, have a more limited capacity for capturing contextual information and dependencies in sequential data like text. Their emphasis on local features further exacerbates their vulnerability to adversarial attacks, leading to the lowest adversarial robustness among the three models.

\section{Summary and Conclusion}
\label{Conc}
This paper introduces a unique investigation into the interplay between explainability and adversarial robustness in hate-speech detection models. Informed by this investigation, we adopt an approach whereby attacks are conducted on widely used hate-speech models, with a focus on exploiting explainable features to reveal vulnerabilities. Our findings provide empirical support for our initial hypothesis, underscoring the potential tradeoff between enhancing model explainability and inadvertently increasing its vulnerability to perturbations, thus compromising adversarial robustness.

Moreover, our study yields compelling evidence of a proportional relationship between explainability and adversarial robustness in hate-speech detection models. This discovery offers valuable insights for the fine-tuning of such models, enabling the establishment of an optimal balance between explainability and adversarial robustness. By leveraging this understanding, we can ensure that the models perform well even under adversarial attacks. 

Our experimentation focuses on the common hate-speech detection models including BERT, LSTM, CNN, as well as a baseline RFC. To gain insights into these models, we employ the LIME (Local Interpretable Model-agnostic Explanations) technique, a widely used approach for explaining machine learning models. LIME is chosen for its ability to approximate the behavior of the models, providing valuable explanations and interpretations. By applying LIME, we aim to enhance our understanding of these hate-speech detection models and shed light on their decision-making processes.

By analyzing the contributions of words from the explainability that was obtained, we are able to target words with high explainability scores and perturb the input to generate adversarial examples, thus assessing the adversarial robustness of each model. This is achieved via the TextAttack tool. Our results demonstrate that models with a higher degree of explainability exhibit increased susceptibility to adversarial attacks. This vulnerability arises from the fact that adversaries can exploit the very features that directly contribute to the model's decision-making process.  

This research emphasizes the significance of considering both explainability and adversarial robustness in the design of hate-speech detection models. By carefully balancing these two aspects, we can create models that are not only accurate and interpretable but also capable of withstanding adversarial attacks that exploit the model's decision-making process. This balanced approach is crucial for ensuring the efficacy, reliability, and practicality of these models across different domains and platforms. This, in turn, leads to enhancements in downstream applications like automated moderation and mediation, ultimately fostering a culture of respectful communication and civil discourse.

\section{Limitations and Future Work}

The current experimental setup has a limitation due to the constraint imposed by the TextAttack tool. This constraint restricts the attack capability to sentences that are shorter than 100 characters. As a result, a considerable number of sentences in certain datasets are excluded from the analysis.

Regarding future work, there is an opportunity to conduct a range of attacks on multiple datasets. The insights gained from these attacks can be leveraged to enhance hate-speech detection models by increasing their Adversarial Robustness. This may involve exploring diverse adversarial techniques, evaluating the models on diverse datasets, and incorporating the lessons learned into the model development process.

\section{Broader Impact and Ethical Considerations}

Our study explores the tradeoff between explainability and adversarial robustness in hate-speech detection models. This research has the potential to deepen our comprehension of the functioning of these models and offer valuable insights for enhancing their design. Additionally, it can contribute to the improvement of downstream tasks like moderation and mediation for promoting ethical discourse, leading to more effective and secure utilization of these models.

The positive impacts of this study include the potential to enhance the performance and security of hate speech detection models, which could lead to more effective and accurate moderation on digital platforms. However, we fully acknowledge the potential for negative outcomes. The adversarial attacks we employ for testing model robustness could, in theory, be misused to weaken the effectiveness of hate speech detection models and propagate harmful content. 

It is also important to note that while we strive for model explainability, there is a risk of over-simplifying complex models, which could lead to misinterpretation where the simplified explanation does not accurately reflect the model's behavior.

To mitigate the potential negative outcomes, we highlight the tradeoffs between explainability and adversarial robustness and emphasize the importance of carefully balancing these two aspects. Our research focuses on understanding the vulnerabilities of these models and exploring ways to improve their resilience against adversarial attacks. Moreover, the insights obtained from this information can be utilized to develop models that are more robust and reliable, ultimately providing a means to foster civil discourse amidst a vast sea of potentially harmful interactions.

In terms of data ethics, our study utilizes public datasets and maintains the privacy and anonymity of the data subjects. This exploration does not involve collection of any new datasets. In the event of future data collection, we will strictly adhere to ethical guidelines.

%\bibliography{NLPProject, custom}

\end{document}